\documentclass[11pt]{article}
\usepackage[preprint]{acl}

\usepackage{times}
\usepackage{latexsym}
\usepackage[T1]{fontenc}
\usepackage[utf8]{inputenc}
\usepackage{microtype}
\usepackage{graphicx}
\usepackage{amsmath}
\usepackage{amssymb}
\usepackage{booktabs}
\usepackage{multirow}
\usepackage{enumitem}
\usepackage{tikz}
\usetikzlibrary{positioning,arrows.meta,shapes.geometric,fit,backgrounds}
\usepackage{xcolor}

\newcommand{\sys}{Profile-Graph Memory}
\newcommand{\sysshort}{ProGraph}
\newcommand{\memhop}{MemHop}

\title{\sys{} for LLM Agents:\\Implicit Cross-Entity Traversal through Narrative Profiles}

\author{Shengtong Zhu \\
  Independent Researcher \\
  \texttt{albertzhushengtong@gmail.com}}

\begin{document}
\maketitle

\begin{abstract}
Long-term memory is essential for LLM agents that interact across sessions, yet current memory benchmarks primarily evaluate single-hop recall, leaving multi-hop association largely unmeasured. We make three contributions. First, we introduce \textbf{\memhop{}}, a multi-hop memory benchmark of 1{,}000 questions at hop depths $1$--$5$ across 10 social-network scenarios, with per-hop evidence annotations. Second, we present \textbf{\sys{}} (\sysshort{}), a two-layer memory architecture combining (i) \emph{profile expansion}---substring-matched traversal of entity names that naturally appear in LLM-written profile narratives, a minimal alternative to explicit knowledge-graph construction---and (ii) \emph{compression residuals}---exact dates, quantities, and named items co-extracted with each profile update at zero extra API cost. Third, a full-grid ablation shows cross-benchmark mechanism specialization: profile expansion drives multi-hop reasoning ($-22.6$pp on \memhop{} when removed) while compression residuals drive precision recall ($-8.6$pp on LoCoMo when not co-extracted), with cross-effects under 3pp within a single architecture. \sysshort{} averages 80.1\% on \memhop{} (matching the FullContext reference) and 78.4\% on LoCoMo (exceeding FullContext by 11.3pp), outperforming Mem0, A-Mem, HippoRAG, and RAG on both. We release \memhop{}, \sysshort{}, and baseline implementations.
\end{abstract}

\section{Introduction}
\label{sec:intro}

As LLM agents persist across many sessions in roles such as personal companions, copilots, and digital life-loggers \cite{memgpt,generative_agents,mem0}, they must remember what users said previously. Existing memory benchmarks (LoCoMo \cite{locomo}; LongMemEval \cite{longmemeval}) are dominated by \emph{precision recall}: asking for a specific date, quantity, or attribute. Yet many valuable agent behaviors require \emph{multi-hop association}: answering ``\emph{what instrument does Alice's roommate play?}'' requires traversing an entity chain Alice~$\to$~Bob~$\to$~piano rather than retrieving a single proposition. How memory systems handle such queries is largely unmeasured.

\paragraph{Why current memory architectures fail multi-hop.}
Consider a personal-assistant agent asked the question above. Three memory paradigms each fail in characteristic ways. \textbf{Atomic-fact stores} \cite{mem0,amem} index dialogue as fine-grained propositions and retrieve by cosine; the query embedding is dominated by ``Alice'' and ``instrument'' and ranks ``\emph{Alice practices guitar}'' first, missing the cross-entity link---atomic methods \emph{lack association}. \textbf{Graph-RAG methods} \cite{hipporag} extract an explicit knowledge graph and run entity-conditioned traversal, but the pipeline must commit, at write time, to which edges the graph will contain---\emph{before any query is seen}---on noisy dialogue where relations are often implicit and surface under different forms across sessions that may fail to merge. A single missing or inconsistently-extracted bridge edge then breaks \emph{every} downstream multi-hop query that needs to traverse it: graph methods \emph{amplify noise} (one-error, many-failures, a pattern atomic stores do not exhibit). \textbf{Concept-profile methods} \cite{ariadnemem} aggregate facts into entity-centric narratives that \emph{contain} cross-entity associations in their text, but standard retrieval surfaces only the entity named in the query---the chain stays buried in narrative without an explicit traversal step---and summarization paraphrases precise details away as profiles grow: profile methods \emph{don't traverse, and lose precision}.

\paragraph{Profile expansion.}
\sysshort{}'s key proposal is to traverse cross-entity chains through the entity names that LLM-written profile narratives \emph{naturally contain}. When an LLM writes a profile of Alice---``\emph{Alice lives with her roommate Bob, who is a multi-instrumentalist}''---the string ``Bob'' appears inside Alice's profile by language alone, without any relation-extraction step or graph construction. At retrieval time we scan each retrieved profile for such mentions and \emph{pull the matched neighbour entities' profiles and residuals into the answer context}, gated by query-embedding relevance. This is cheap to compute, robust to extraction noise (because no graph is constructed in the first place), and---crucially---makes no write-time commitment about which edges the future will need.

\paragraph{Compression residuals.}
Narrative profiles carry association well but paraphrase precise details (dates, quantities, named items) away. We complement the profile with a second layer of \emph{compression residuals}: short atomic facts that the LLM is asked to identify \emph{alongside each profile update}, in the same write-time call. This adds no extra LLM round trip---only tens to a few hundred output tokens per entity---making the precision layer essentially free.

\paragraph{Contributions.}
\textbf{(1) \memhop{} benchmark} (§\ref{sec:benchmark}): 1{,}000 multi-hop questions at hop depths $K{=}1$--$5$ (the number of entity-bridging steps to the answer) over 10 social-network scenarios, with per-hop decomposition and gold evidence. \textbf{(2) \sys{} architecture} (§\ref{sec:method}): combines profile expansion (§\ref{sec:method:read}) and compression residuals (§\ref{sec:method:write}) in a single-call write path. \textbf{(3) Mechanism specialization} (§\ref{sec:experiments:factorial}): expansion carries multi-hop ($-22.6$pp on \memhop{} when removed); extraction-time residuals carry precision ($-8.6$pp on LoCoMo when removed)---the same architecture handles both query types without task-specific switching.

\section{Related Work}
\label{sec:related}

\paragraph{Multi-hop QA: static vs.\ conversational.}
Multi-hop QA over \emph{static} corpora has well-developed benchmarks (HotpotQA \cite{hotpotqa}, MuSiQue \cite{musique}, 2WikiMultiHopQA \cite{2wikimultihop}) and methods that build explicit structure---HippoRAG \cite{hipporag} extracts an entity-relation knowledge graph and traverses it with personalized PageRank (PPR), GraphRAG \cite{graphrag} clusters entities into communities for hierarchical retrieval, and LightRAG \cite{lightrag} couples dual-level retrieval with a small graph index. These methods report strong numbers on clean corpora where extraction errors are rare. \emph{Conversational} memory benchmarks (LoCoMo \cite{locomo}, LongMemEval \cite{longmemeval}) instead emphasize precision recall: a date, a quantity, a preference. The conversational $\times$ multi-hop intersection is largely unmeasured. Transferring HippoRAG to \memhop{} we find its PPR traversal recovers only a modest signal over chunk RAG (57.2 vs.\ 55.9), trailing \sysshort{} by 22.9pp---explicit KG construction is dominated in this regime by what LLM-written profile narratives already encode.

\paragraph{Atomic-fact, profile, and hybrid memory.}
Single-layer atomic-fact systems---Mem0 \cite{mem0}, A-Mem \cite{amem} (with Zettelkasten write-time linking), MemoryBank \cite{memorybank}, AdaMem \cite{adamem}, SimpleMem \cite{simplemem}---decompose dialogue into fine-grained propositions retrieved by embedding similarity. These excel at single-hop lookup but, as our \memhop{} results confirm, collapse on $K{\geq}2$ chains where the answer entity is semantically distant from the query \cite{hipporag}. Single-layer profile-centric methods (AriadneMem \cite{ariadnemem}) aggregate facts into narrative summaries that naturally encode associations but paraphrase precision-critical details away. Concurrent contemporaneous work combines atomic and profile layers via two-stage extraction pipelines \cite{trimem,memori,hindsight}, but none are designed for or evaluated on $K$-hop chains. \sysshort{} differs along three combined axes: (i) profile and compression residuals are co-extracted in a \emph{single} LLM call (zero extra API cost rather than 2$\times$ write), (ii) multi-hop traversal emerges from substring co-mention inside narrative profiles rather than any explicit graph structure, and (iii) we evaluate explicitly on multi-hop chains at hop depths $K{=}1$--$5$.

\section{\sys{}}
\label{sec:method}

\begin{figure*}[t]
  \centering
  \definecolor{boxFill}{HTML}{F5F7FA}
  \definecolor{boxStroke}{HTML}{4A5568}
  \definecolor{accentLLM}{HTML}{D97706}
  \definecolor{accentStore}{HTML}{2563EB}
  \definecolor{accentEmbed}{HTML}{059669}
  \definecolor{captionGray}{HTML}{718096}
  \newcommand{\subt}[1]{\\{\scriptsize\color{boxStroke!90} #1}}
  \begin{tikzpicture}[
    scale=0.82, transform shape,
    font=\small,
    node distance=0.3cm and 0.7cm,
    base/.style={rectangle, rounded corners=2.5pt, draw=boxStroke, line width=0.5pt, fill=boxFill, minimum height=0.7cm, align=center, inner sep=3pt},
    llm/.style={base, draw=accentLLM, line width=0.8pt, fill=accentLLM!8},
    embed/.style={base, draw=accentEmbed, line width=0.8pt, fill=accentEmbed!8},
    store/.style={rectangle, rounded corners=4pt, draw=accentStore, line width=0.7pt, fill=accentStore!6, align=center, minimum width=3.1cm, minimum height=0.9cm, inner sep=3pt},
    arr/.style={-{Stealth[length=4pt]}, line width=0.5pt, draw=boxStroke},
    arrDashed/.style={-{Stealth[length=4pt]}, line width=0.5pt, draw=accentStore, dashed},
    header/.style={font=\small\bfseries},
    capt/.style={font=\scriptsize\itshape, color=captionGray}
  ]
    \node[header] (writeH) at (-5.5, 4.1) {Write path \textnormal{\scriptsize (per session)}};
    \node[header] (readH)  at ( 5.5, 4.1) {Read path \textnormal{\scriptsize (per query)}};

    \node[base, minimum width=4.0cm] (session) at (-5.5, 3.2) {Session $c_t$ \subt{speaker + turns + timestamp}};
    \node[llm, minimum width=4.0cm, below=of session] (step1) {Step 1: Entity identification \subt{1 LLM call}};
    \node[llm, minimum width=4.0cm, below=of step1] (step2) {Step 2: For each entity $e_i$, \\ \textbf{joint profile + residual co-extract} \subt{1 LLM call per entity}};

    \draw[arr] (session) -- (step1);
    \draw[arr] (step1) -- (step2);

    \node[store] (profs) at (0, 2.0) {\textbf{Profiles}\\ $\{e_i \mapsto p_i\}$ \subt{embedded}};
    \node[store, below=0.3cm of profs] (resids) {\textbf{Residuals}\\ $\{e_i \mapsto [r_1,\dots,r_m]\}$ \subt{entity-bound, embedded}};

    \draw[arr] (step2.east) to[out=0, in=180] (profs.west);
    \draw[arr] (step2.east) to[out=0, in=180] (resids.west);

    \node[base, minimum width=4.0cm] (query) at (5.5, 3.2) {Query $q$};
    \node[embed, minimum width=4.0cm, below=of query] (s1) {Stage 1: Profile retrieval \subt{score every profile; top-$M$ by $\cos(\mathbf{v}_q,\mathbf{v}_{p_i})$ + name boost}};
    \node[embed, minimum width=4.0cm, below=of s1] (s2) {Stage 2: Relevance-gated expansion \subt{scan profiles for entity names; gate by $\cos > \tau_{\text{expand}}$}};
    \node[embed, minimum width=4.0cm, below=of s2] (s3) {Stage 3: Residual-augmented context \subt{profile + top query-relevant residuals (capped per entity)}};
    \node[llm, minimum width=4.0cm, below=of s3] (ans) {Answer generation \subt{1 LLM call}};

    \draw[arr] (query) -- (s1);
    \draw[arr] (s1) -- (s2);
    \draw[arr] (s2) -- (s3);
    \draw[arr] (s3) -- (ans);

    \draw[arrDashed] (profs.east) to[out=0, in=180] (s1.west);
    \draw[arrDashed] (profs.east) to[out=0, in=180] (s2.west);
    \draw[arrDashed] (resids.east) to[out=0, in=180] (s3.west);

    \node[capt, text width=4.5cm, align=center] at (-5.5, -3.1) {Single LLM call produces \emph{both} profile and residuals---zero extra API cost.};
    \node[capt, text width=4.5cm, align=center] at ( 5.5, -3.1) {Pure-embedding retrieval; only 1 LLM call (answer generation).};

    \begin{scope}[shift={(0,-3.1)}]
      \node[llm, minimum width=0.5cm, minimum height=0.3cm] (L1) at (-2.6,0) {};
      \node[anchor=west, font=\scriptsize] at (L1.east) {~LLM call};
      \node[embed, minimum width=0.5cm, minimum height=0.3cm] (L2) at (-0.6,0) {};
      \node[anchor=west, font=\scriptsize] at (L2.east) {~pure-embedding};
      \node[store, minimum width=0.5cm, minimum height=0.3cm] (L3) at (2.6,0) {};
      \node[anchor=west, font=\scriptsize] at (L3.east) {~store};
    \end{scope}
  \end{tikzpicture}
  \caption{\sys{} architecture. \textbf{Write} (left): each session is parsed for entities, and for each entity a \emph{single} LLM call produces the updated profile and the residuals it could not preserve, at zero extra API cost. \textbf{Read} (right): three pure-embedding stages (profile retrieval $\to$ relevance-gated expansion over entity names in profile text $\to$ residual-augmented context assembly) feed one answer-generation call. Residuals are entity-bound, never indexed globally.}
  \label{fig:architecture}
\end{figure*}
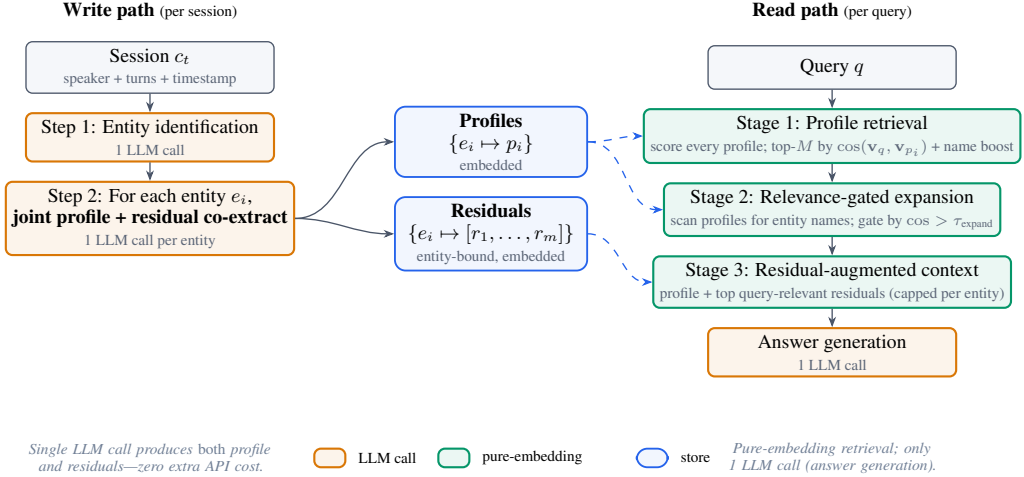

\sysshort{} maintains two layers per entity $e_i$: an \emph{entity profile} $p_i$ (narrative summary supporting association) and a set of \emph{compression residuals} $R_i$ (precision-critical facts the summary discards). Both are co-extracted in a single LLM call. Figure~\ref{fig:architecture} summarizes the architecture.

\subsection{Write path: joint co-extraction}
\label{sec:method:write}
Each new session $c_t$ is processed in three steps. \textbf{(1) Entity identification}: an LLM call returns the entity names mentioned in the session; previously tracked ones are matched via a name registry. \textbf{(2) Profile + residual co-extraction}: for each entity a single LLM call takes the current profile $p_i$ and the relevant turns of $c_t$ and outputs an updated profile $p_i'$ \emph{and} a set of residuals $R_i'$. The prompt asks for residuals in seven categories: exact dates/times, numbers and quantities, proper names (titles, places), identity markers, enumerated lists, cross-entity facts (recommendations, shared experiences), and state transitions (full prompts released with the code). Crucially, this is not generic fact extraction: the LLM is asked for \emph{what the summary would lose}, making residuals naturally complementary to the profile. \emph{Why not just write a more complete profile?} Empirically, narrative form structurally paraphrases precise dates and quantities into smoother phrasing even when asked for completeness---collapsing the two layers into one costs $-8.6$pp on LoCoMo (§\ref{sec:experiments:factorial}). The two-layer design also pays off at retrieval: short atomic residuals match precision queries with sharper cosine than long narratives do. Each residual is self-contained and includes the entity name. \textbf{(3) Storage and dedup}: profile embeddings and per-entity residual embeddings are stored; a residual is discarded if its cosine similarity to any existing residual of the same entity exceeds $\tau_{\text{dedup}}{=}0.9$.

\subsection{Read path}
\label{sec:method:read}
Given query $q$, retrieval is three pure-embedding stages followed by one answer-generation LLM call.

\textbf{Stage~1 (profile retrieval):} embed $q$ and score every profile by cosine; entities explicitly named in $q$ receive a name-boost $\beta{=}0.3$. The top-$M{=}5$ entities form the initial \emph{selection set} $S$---the set of entities whose profiles and residuals will feed the answer LLM after Stage~2 expansion grows $S$ with relevant neighbours.

\textbf{Stage~2 (relevance-gated expansion):} multi-hop queries typically have answers in a different entity than the one named---``\emph{what instrument does Alice's roommate play?}'' embeds close to Alice but the answer is in Bob. We treat each profile's narrative as an \emph{implicit entity graph}: because the LLM writes Alice's profile as ``\emph{Alice lives with her roommate Bob, who is a multi-instrumentalist}'' (using the same example as §\ref{sec:intro}), the string \texttt{Bob} appears inside and constitutes an implicit edge Alice~$\to$~Bob. Starting from the Stage~1 selection set $S$, we iterate the expansion at most $H_{\max}{=}5$ times (sensitivity analysis in Appendix~\ref{app:hmax}). $H_{\max}$ is a hyperparameter capping how many expansion hops Stage~2 may take (an \emph{algorithm-side} budget, distinct from a question's hop depth $K$ which is a property of the benchmark). At each iteration we scan profiles in $S$ for already-registered entity names (Appendix~\ref{app:string-match}), score each candidate $e_j$ by $s_j = \cos(\mathbf{v}_q, \mathbf{v}_{p_j})$, and add the candidates to $S$ if $\max_j s_j > \tau_{\text{expand}}{=}0.2$; otherwise we stop. The relevance gate makes expansion adaptive: dense neighbourhoods do not cause fan-out because each hop must clear the gate.

\textbf{Stage~3 (residual-augmented context):} for each selected entity $e_i \in S$, \emph{independently}: the profile $p_i$ is presented first; then, from the residuals $R_i$ accumulated for $e_i$ across past write-path calls, we rank by $\cos(\mathbf{v}_q, \mathbf{v}_{r_j})$ and append the top ones (subject to a small per-entity cap) labeled as ``Verified Facts.'' Rankings are not shared across entities---each $e_i$ gets its own top slice---so a residual-rich entity cannot crowd out a residual-poor one. Residuals of unselected entities never appear. The answer prompt instructs the LLM that verified facts override the profile in case of conflict.

\subsection{Worked example: Alice's roommate}
\label{sec:method:example}
We trace the read path on the running example \emph{``what instrument does Alice's roommate play?''} ($K{=}2$; gold: piano).

\textbf{Stage~1.} Alice's profile ranks top-1 by cosine: the query names her, and the name-boost $\beta{=}0.3$ further lifts her score. The remaining top-$M$ slots go to entities with general lexical overlap to ``instrument''. Bob---the answer-bearing entity---does \emph{not} rank top-$M$ directly because the query does not name him.

\textbf{Stage~2.} Scanning profiles in $S$ for registered entity names finds the string ``Bob'' inside Alice's narrative ``\emph{Alice lives with her roommate Bob, who is a multi-instrumentalist}''. Bob's profile is scored against the query; the score clears $\tau_{\text{expand}}{=}0.2$, and Bob joins $S$. Other co-mentioned entities whose profiles do not clear the gate are not added---the relevance gate prevents fan-out into the dense neighbourhood.

\textbf{Stage~3.} For each entity in $S$, residuals are ranked by cosine to the query. Bob's residual ``\emph{plays piano, drums, and bass guitar regularly}'' ranks top-1 (it shares instrument-related terms with the query) and surfaces alongside his profile. The answer LLM receives \{Alice's profile, Bob's profile, the top residuals retrieved per entity\} and answers correctly. By contrast, Mem0 answers ``guitar'' (incorrect): its cosine-only retrieval over atomic facts surfaces the closest match ``\emph{Alice plays acoustic guitar}'' rather than traversing to Bob's profile---atomic stores cannot cross the implicit bridge from Alice to Bob.

\subsection{Cost}
\begin{table}[t]
\centering
\footnotesize
\setlength{\tabcolsep}{4pt}
\begin{tabular}{lcc}
\toprule
\textbf{System} & \textbf{Write}$^\dagger$ & \textbf{Read} \\
 & (per entity) & (per query) \\
\midrule
Mem0          & 2 (extract + dedup)   & 1 \\
A-Mem         & 2 (extract + evolve)  & 1 \\
Profile-only  & 1 (update)            & 1 \\
\sysshort{} (ours) & \textbf{1} (co-extract) & \textbf{1} \\
\bottomrule
\end{tabular}
\caption{LLM calls per write operation (per entity) and per read query. $^\dagger$Excludes a one-time $O(1)$ per-session call to identify the entity set, shared by all profile-based systems.}
\label{tab:cost}
\end{table}

Per session \sysshort{} makes $1 + E$ LLM calls ($E$ = average entities per session), the same as a profile-only memory (Table~\ref{tab:cost}); the residual output adds tens to a few hundred tokens to an already-scheduled call. Read cost is one LLM call (answer generation); the three retrieval stages are pure embedding operations. A naive design that extracts residuals in a separate LLM call would double the per-entity write cost without any retrieval benefit, since the LLM has already loaded the relevant context to write the profile.

\section{\memhop{} Benchmark}
\label{sec:benchmark}

Existing benchmarks split the conversational and multi-hop dimensions (§\ref{sec:related}); \memhop{} fills this gap by evaluating end-to-end---from conversation ingestion through extraction to multi-hop answer generation over an entity bridge graph.

\paragraph{Scenario design.}
Each scenario defines a social network of 4--5 characters connected by diverse relationship types (roommates, colleagues, family, mentors). Across 10 scenarios we have 46 characters, 74 relationships of 63 distinct types, and 920 grounded observations (attributes, preferences, experiences). Observations are embedded into 200 natural dialogue sessions (${\sim}2{,}500$ turns total, ${\sim}12$ turns per session) between a user and an assistant, simulating diary-style interaction. The per-scenario depth profile is fixed: 25 questions at $K{=}1$, 25 at $K{=}2$, 20 at $K{=}3$, 18 at $K{=}4$, 12 at $K{=}5$ (100 per scenario, 1{,}000 in total).

\paragraph{Entity bridge chain method.}
We generate $K$-hop questions by sampling a chain of $K{+}1$ entities linked by $K$ relationships, $e_1 \to e_2 \to \cdots \to e_{K+1}$, from the underlying social graph (each consecutive pair $(e_i, e_{i+1})$ shares a documented relationship). The question starts from $e_1$ and asks about an attribute of $e_{K+1}$. Each question records (i) a \emph{decomposition}---$K$ sub-questions $\{q_1, \ldots, q_K\}$, one per chain edge, derived programmatically from the sampled chain ($q_i$ resolves $e_i \to e_{i+1}$)---and (ii) \emph{gold evidence}: the schema observations that document each relationship. The natural-language composite question is then LLM-rephrased over the chain. Because decomposition and evidence are derived (not LLM-generated), per-hop annotations are exact and enable fine-grained failure analysis (which hop broke?), used in Appendix~\ref{app:error-tab} to categorize \sysshort{}'s errors.

\paragraph{Example ($K{=}3$).}
\emph{Q: What is the favorite cuisine of the person who taught Alice's roommate to cook?}---chain: Alice $\xrightarrow{\text{roommate}}$ Bob $\xrightarrow{\text{cooking teacher}}$ Carol $\xrightarrow{\text{fav.\ cuisine}}$ Italian. Decomposition: $q_1$: \emph{Alice's roommate?} $\to$ Bob; $q_2$: \emph{Who taught Bob to cook?} $\to$ Carol; $q_3$: \emph{Carol's favorite cuisine?} $\to$ Italian.

\paragraph{Quality assurance.}
The 1{,}000 questions pass five rounds of automated validation: (1) structural integrity (valid JSON, required fields); (2) evidence grounding (each sub-answer supported by the cited observation); (3) answer consistency (final answer matches chain endpoint); (4) textual quality (length, naturalness); (5) deduplication (no near-duplicate questions). All construction prompts and pipeline code are released. Additional per-scenario statistics are in Appendix~\ref{app:memhop-construction}.

\section{Experiments}
\label{sec:experiments}

\paragraph{Setup.}
We evaluate on \memhop{} ($n{=}1{,}000$) and LoCoMo \cite{locomo} (10 conversations, $n{=}1{,}986$ QAs averaging $\sim$199/conv, range 105--260). All systems use GPT-4o-mini for LLM calls and all-MiniLM-L6-v2 \cite{minilm} for embeddings. Baselines: \textbf{Oracle} (gold evidence directly, \memhop{} only; isolates LLM reasoning from retrieval), \textbf{FullContext} (entire conversation as one LLM input; bounds the LLM's long-context ceiling), \textbf{Mem0} \cite{mem0}, \textbf{A-Mem} \cite{amem}, \textbf{HippoRAG} \cite{hipporag}, \textbf{RAG} (raw 4-turn chunks); all implementations follow the authors' released code (Appendix~\ref{app:baselines}).\footnote{We use HippoRAG as the representative graph-RAG baseline; GraphRAG \cite{graphrag} and LightRAG \cite{lightrag} are designed for static document corpora and require non-trivial adaptation to multi-session dialogue, which we leave to future work.}

\paragraph{Evaluation.}
Primary metric is LLM-as-Judge accuracy with GPT-4o-mini. \memhop{} uses a single judge (answers are unambiguous attribute values). On LoCoMo's overall results (Table~\ref{tab:locomo}) we report two judges: a \emph{strict} judge that requires near-exact semantic match (e.g., ``\emph{May 7, 2023}'' matches ``\emph{5/7/23}'' but not ``\emph{around May 2023}''), and the more permissive \emph{LightMem} judge \cite{lightmem} that additionally accepts paraphrases and approximate temporal/numeric matches. LightMem is adopted as primary because it is the increasingly common convention in recent memory papers (Mem0~\citep{mem0}, EverMemOS~\citep{evermemos}, LightMem itself), enabling cross-paper comparison. Per-category and ablation analyses use LightMem alone for compactness; strict scores for every cell are in the released artifacts. LoCoMo's adversarial cat-5 questions are designed to have no answer in the conversation; the judge marks a system correct if it refuses (e.g., outputs ``\emph{I don't know}'' or ``\emph{the conversation doesn't say}'') and incorrect if it hallucinates a content answer. We also report F1 for prior-work comparability, while flagging that it penalizes correct verbose answers (see Table~\ref{tab:locomo} caption). Sample sizes ($n{\geq}1{,}000$) put 95\% percentile bootstrap CIs at ${\pm}2$--$3$pp; we omit per-cell CIs to keep tables compact and explicitly flag the one cell where the gap falls inside the CI band (\sysshort{} vs.\ FullContext on \memhop{}, finding 1 below).

\subsection{Main results}
\label{sec:exp:main}

\begin{table}[t]
\centering
\footnotesize
\setlength{\tabcolsep}{4pt}
\begin{tabular}{lcccccc}
\toprule
\textbf{System} & $K{=}1$ & $K{=}2$ & $K{=}3$ & $K{=}4$ & $K{=}5$ & \textbf{Avg} \\
\midrule
\textit{Oracle}      & 97.6 & 78.8 & 84.0 & 87.2 & 88.3 & 87.2 \\
\textit{FullCtx}     & 91.6 & 76.0 & 75.5 & 77.2 & 72.5 & 79.6 \\
\midrule
\sysshort{}          & \textbf{91.6} & \textbf{74.8} & \textbf{81.0} & \textbf{76.1} & \textbf{71.7} & \textbf{80.1} \\
HippoRAG             & 84.4 & 47.2 & 45.5 & 50.6 & 50.8 & 57.2 \\
RAG                  & 83.6 & 47.2 & 42.0 & 47.2 & 52.5 & 55.9 \\
A-Mem                & 68.8 & 34.0 & 33.5 & 28.9 & 36.7 & 42.0 \\
Mem0                 & 54.0 & 28.4 & 35.5 & 32.2 & 32.5 & 37.4 \\
\bottomrule
\end{tabular}
\caption{Accuracy (\%) on \memhop{} by hop depth $K$ (10 scenarios, $n{=}1{,}000$). Bold = best memory system per column.}
\label{tab:memhop}
\end{table}

\paragraph{\memhop{}: \sysshort{} matches FullContext, atomic and graph-RAG collapse.}
Table~\ref{tab:memhop} reports \memhop{} results. \textbf{(1)} \sysshort{} at 80.1\% essentially ties FullContext (79.6\%; 0.5pp is within sampling noise) and trails Oracle (87.2) by 7.1pp. The Oracle--FullContext gap (7.6pp) is a long-context tax that any retrieval system can in principle close; \sysshort{} has closed it. \textbf{(2)} The cliff at $K{=}2$ is universal among atomic/graph-RAG methods, with drops of $25$--$37$pp (Table~\ref{tab:memhop}). The cliff occurs because $K{\geq}2$ questions need to traverse an entity bridge---precisely where cosine similarity fails. \sysshort{}'s $K{=}2$ drop ($91.6 \to 74.8 = 16.8$pp) is roughly half the magnitude of the atomic/graph-RAG cliff and matches Oracle's $18.8$pp drop, because entity profiles encode cross-entity relationships within narrative text: a single retrieval surfaces the multi-hop chain. Both \sysshort{} and Oracle exhibit a non-monotonic dip at $K{=}2$ followed by recovery at $K{=}3$ (\sysshort{}: $74.8 \to 81.0$; Oracle: $78.8 \to 84.0$), a pattern absent for FullContext, RAG, HippoRAG, and A-Mem; we discuss this cross-system anomaly in Appendix~\ref{app:error-tab}. \textbf{(3)} HippoRAG's explicit KG construction provides only a 1.3pp gain over chunk RAG, suggesting that on noisy conversational extractions the structural machinery is dominated by what the narrative already contains: \sysshort{}'s language-level traversal outperforms HippoRAG by 22.9pp.

\begin{table}[t]
\centering
\small
\begin{tabular}{lccc}
\toprule
\textbf{System} & \textbf{LightMem} & \textbf{Strict} & \textbf{F1} \\
\midrule
\textit{FullContext} & 67.1 & 61.0 & 0.231 \\
\midrule
\sysshort{}          & \textbf{78.4} & \textbf{71.3} & 0.241 \\
HippoRAG             & 64.1 & 57.6 & 0.218 \\
RAG                  & 60.1 & 53.3 & 0.204 \\
A-Mem                & 55.6 & 47.6 & 0.238 \\
Mem0                 & 52.7 & 47.8 & \textbf{0.283} \\
\bottomrule
\end{tabular}
\caption{LoCoMo accuracy (\%) across all 10 conversations ($n{=}1{,}986$). Bold = best memory system per column. F1 is token-overlap and disadvantages verbose answers (Mem0's ${\sim}3.9$-word vs.\ \sysshort{}'s ${\sim}10.9$-word responses against ${\sim}4$-word gold), motivating LLM-as-Judge for the accuracy columns.}
\label{tab:locomo}
\end{table}

\paragraph{LoCoMo: \sysshort{} exceeds FullContext.}
On LoCoMo (Table~\ref{tab:locomo}), \sysshort{} reaches 78.4\% (LightMem) / 71.3\% (strict), exceeding the FullContext reference by 11.3 / 10.3pp---selective retrieval filters the irrelevant context that otherwise dilutes long-context LLM attention. \sysshort{} leads HippoRAG by 14.3pp here, mirroring the 22.9pp lead on \memhop{}---the residual + implicit-graph combination dominates the explicit-graph + raw-chunk approach across both regimes.

\begin{table*}[t]
\centering
\small
\begin{tabular}{lccccc}
\toprule
\textbf{System} & \textbf{Multi-hop} & \textbf{Temporal} & \textbf{Open-dom.} & \textbf{Single-hop} & \textbf{Adversarial} \\
\midrule
\textit{FullContext} & \textbf{82.3} & 48.9 & \textbf{64.6} & \textbf{94.4} & 20.6 \\
\midrule
\sysshort{}          & 78.7 & \textbf{77.6} & 43.8 & 83.9 & \textbf{74.9} \\
HippoRAG             & 72.0 & 44.2 & 62.5 & 85.4 & 34.8 \\
RAG                  & 66.0 & 38.0 & 57.3 & 79.8 & 37.9 \\
A-Mem                & 38.7 & 52.3 & 25.0 & 58.9 & 67.7 \\
Mem0                 & 50.0 & 60.4 & 33.3 & 57.3 & 45.1 \\
\bottomrule
\end{tabular}
\caption{Per-category LoCoMo accuracy (\%) under the LightMem judge ($n{=}1{,}986$). Bold = best per column.}
\label{tab:locomo_cat}
\end{table*}

\paragraph{Per-category LoCoMo.}
Table~\ref{tab:locomo_cat} reports per-category accuracy under the LightMem judge.\footnote{LoCoMo's released JSON encodes the five categories as multi-hop / temporal / open-domain / single-hop / adversarial; this differs from the textual ordering in the LoCoMo paper §4.1. We use the encoding present in the released JSON.} \sysshort{} achieves its largest gains on \emph{temporal} (+28.7pp over FullContext), where residuals preserve exact dates resolved from relative expressions against session timestamps (following TSM \cite{tsm}), and on \emph{adversarial} (+54.3pp), where two factors compound. (i) \emph{Retrieval granularity}: profile-based or atomic systems (\sysshort{} 74.9, A-Mem 67.7, Mem0 45.1) hand the answer LLM short, semantically focused units---when none of them addresses the query the LLM sees the absence cleanly and refuses; chunk-based retrieval (HippoRAG 34.8, RAG 37.9) and whole-conversation context (FullContext 20.6) instead supply peripheral text that the LLM pattern-matches into a confident wrong answer. (ii) \emph{Relevance gating}: on top of the granularity effect, \sysshort{}'s gate $\tau_{\text{expand}}{=}0.2$ stops Stage~2 expansion when no neighbour clears it, which we believe accounts for the further $+7.2$pp \sysshort{} holds over A-Mem. \emph{Open-domain} (43.8 vs.\ FullContext 64.6 / HippoRAG 62.5) is the headline weakness: synthesizing soft preferences across many peripheral facts demands the kind of verbatim retention that profile summarization paraphrases away.

\subsection{Ablation: mechanism specialization}
\label{sec:experiments:factorial}

We isolate three switches: \textbf{Expansion} (Stage~2 relevance-gated expansion on/off), \textbf{Read-time residuals} (residuals visible to the answer LLM at Stage~3 on/off; the extraction prompt is unchanged), and \textbf{Write-time residuals} (residuals produced by the extraction prompt on/off; off means the LLM is given a profile-only prompt at write time). Each cell averages over all 10 \memhop{} scenarios and all 10 LoCoMo conversations.

\begin{table}[t]
\centering
\footnotesize
\setlength{\tabcolsep}{3pt}
\begin{tabular}{ccc cc cc}
\toprule
\multicolumn{3}{c}{\textbf{Switches}} & \multicolumn{2}{c}{\textbf{MemHop}} & \multicolumn{2}{c}{\textbf{LoCoMo}} \\
\cmidrule(lr){1-3} \cmidrule(lr){4-5} \cmidrule(lr){6-7}
Exp. & R-time & W-time & Acc. & $\Delta$ & Acc. & $\Delta$ \\
\midrule
\checkmark & \checkmark & \checkmark & \textbf{80.1} & --- & \textbf{71.3} & --- \\
\checkmark &            & \checkmark & 79.2 & $-0.9$ & 70.4 & $-0.9$ \\
\checkmark & \checkmark &            & 77.2 & $-2.9$ & \textbf{62.7} & $\mathbf{-8.6}$ \\
           & \checkmark & \checkmark & \textbf{57.5} & $\mathbf{-22.6}$ & 70.6 & $-0.7$ \\
           & \checkmark &            & 57.5 & $-22.6$ & 69.9 & $-1.4$ \\
\bottomrule
\end{tabular}
\caption{Ablation (\%) on \memhop{} and LoCoMo. \textbf{Switches}: \textbf{Exp.}~=~Stage~2 expansion; \textbf{R-time}~=~residuals shown to the answer LLM at read time; \textbf{W-time}~=~residuals produced by the extraction prompt at write time.}
\label{tab:ablation}
\end{table}

\paragraph{Central finding: cross-benchmark mechanism specialization.}
On \memhop{}, disabling expansion costs 22.6pp while disabling write-time residual extraction costs 2.9pp; on LoCoMo the relative magnitudes invert (expansion: $-0.7$pp; write-time residuals: $-8.6$pp). The two benchmarks are each dominated by a different mechanism by an order of magnitude, and the two mechanisms coexist in one architecture without task-specific routing. \memhop{}'s questions traverse $K{\geq}2$ entity chains where Stage~2 expansion is the key bridge; LoCoMo demands precise attributes that the profile narrative would otherwise paraphrase away. For instance, on the date query \emph{``on what date did Caroline take the pottery class?''}, Caroline's profile compresses the event to ``\emph{passionate amateur potter since 2023}'' but her residual ``\emph{May 7, 2023}'' is ranked top-1 at Stage~3 and surfaces the precise date in the answer context. A per-$K$ decomposition of the expansion gain, including a counterintuitive regression at $K{=}5$, is in Appendix~\ref{app:hmax}.

\paragraph{Methodological note.}
The read-time switch (hide residuals from the answer LLM but keep them in store) makes residuals look near-inert ($-0.9$pp on both benchmarks) because the joint extraction prompt lets profiles absorb the same facts. Only the write-time switch (profile-only extraction prompt) reveals what residuals contribute when the profile must stand alone---an order of magnitude larger on LoCoMo.

\subsection{Accuracy vs.\ hop depth}
\label{sec:exp:khop}

\begin{figure}[t]
  \centering
  \includegraphics[width=\columnwidth]{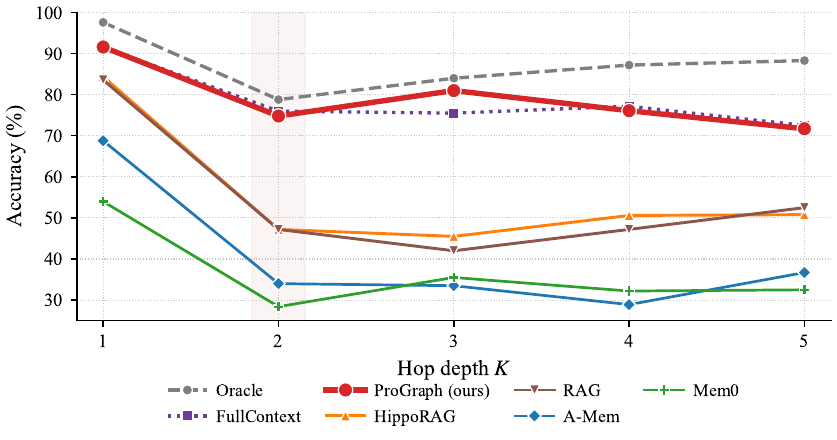}
  \caption{Accuracy (\%) vs.\ hop depth $K$ on \memhop{} (10-scenario average, $n{=}1{,}000$).}
  \label{fig:accuracy_vs_k}
\end{figure}

Figure~\ref{fig:accuracy_vs_k} visualizes the per-$K$ trends. Two regimes are immediately visible: \sysshort{} joins Oracle and FullContext in a high band that stays above 70\% for every $K$, while the four cosine-only memory and retrieval baselines (HippoRAG, RAG, A-Mem, Mem0) collapse to the 28--55\% range from $K{=}2$ onwards and remain there. \sysshort{} is the only memory architecture in the high band; the ${\sim}20$-percentage-point separation between the bands at $K{\geq}2$ is the precision--association trade-off in effect.

\paragraph{Failure modes.}
\sysshort{}'s 19.9\% of \memhop{} misses split 5.5\% / 81.9\% / 12.6\% across three categories: single-hop precision losses (profile compressed the asked-for attribute), multi-hop traversal failures (expansion terminated at a semantically neutral entity, or the LLM slipped on surface-form-confused entities), and LLM refusals (even with relevant context, the answer LLM declined). Refusals dominate the small single-hop failure set ($\sim$48\% of $K{=}1$ misses vs.\ 12.6\% overall); multi-hop traversal failures concentrate at $K{=}2$ and persist across $K{\geq}3$, indicating the difficulty is structural rather than monotonic in chain length (Appendix~\ref{app:error-tab}).

\section{Discussion}
\label{sec:discussion}

\paragraph{Why explicit KG construction underperforms here.}
On static corpora, HippoRAG \cite{hipporag} and GraphRAG \cite{graphrag} report strong multi-hop numbers because the input is curated and extraction errors are rare. The conversational regime inverts this: noisy multi-session dialogue makes the write-time graph commitment (§\ref{sec:intro}) brittle, and the 22.9pp \memhop{} gap between \sysshort{} and a faithful HippoRAG implementation (§\ref{sec:exp:main}) quantifies what this costs.

\paragraph{Why one architecture handles both regimes.}
The full-grid ablation (Table~\ref{tab:ablation}) shows the two layers are decoupled by mechanism: expansion carries multi-hop, write-time residuals carry precision, with cross-effects ${\leq}3$pp. Because each layer's cost is incremental---expansion adds only cosine scores, residuals add only output tokens to an already-scheduled LLM call---combining them does not require choosing per query. This is the practical payoff of mechanism specialization: a unified store with task-agnostic routing.

\section{Conclusion}
\label{sec:conclusion}

We introduced \memhop{}, a multi-hop conversational-memory benchmark; \sys{} (\sysshort{}), a two-layer architecture co-extracting profiles and compression residuals at zero extra API cost; and showed through a full-grid ablation that expansion and residual extraction serve specialized roles (expansion drives multi-hop; residuals drive precision) within a unified architecture. \sysshort{} matches FullContext on \memhop{} and exceeds it on LoCoMo, outperforming atomic and graph-RAG baselines by large margins on both. We release \memhop{}, \sysshort{}, all baseline implementations, and our reproducibility chain at \url{https://github.com/ShengtongZhu/ProGraph}.

\section*{Limitations}

\paragraph{Open-domain query weakness.}
\sysshort{} underperforms on LoCoMo cat~3 (open-domain), reaching 43.8 (LightMem) / 28.1 (strict). The canonical failure: on \emph{``what would Alice enjoy as a birthday gift?''}, where the gold answer requires synthesizing peripheral facts (an aging cat, a gardening question, a French-press joke), Mem0 retrieves the gardening fact verbatim and answers correctly, while \sysshort{}'s Alice profile compresses these into ``\emph{enjoys various hobbies}'' and answers incorrectly. Profile summarization paraphrases such peripheral preferences away, and residuals---being precision-targeted---do not recover them; closing this gap likely requires a third storage layer for open-ended synthesis.

\paragraph{Benchmark scope.}
We evaluate on LoCoMo and \memhop{}---both directly test cross-session multi-hop and $K$-hop chains. LongMemEval \cite{longmemeval}, where 96\% of context is distractor sessions, primarily measures retrieval robustness rather than memory architecture per se; our cat-5 adversarial evaluation (\sysshort{} 74.9 vs.\ Mem0 45.1) covers similar territory, and a pilot run confirms our pipeline produces sensible LongMemEval predictions. A full LongMemEval grid would require ${\sim}28$ wallclock hours under our shared inference budget and is left to future work.

\paragraph{Embedder and judge.}
We use all-MiniLM-L6-v2 \cite{minilm} embeddings and GPT-4o-mini \cite{gpt4omini} as both LLM and judge; SOTA embeddings would likely raise absolute numbers but \sysshort{}'s string-matched expansion is embedding-agnostic, and LLM-as-judge biases are symmetric across systems. Our Oracle prompt is deliberately conservative for cross-system symmetry; a more lenient ``commit-to-answer'' Oracle reaches 95.5\% (8.3pp above the reported 87.2\%), indicating substantial retrieval-precision headroom before the answer-LLM becomes the bottleneck.

\paragraph{Hyperparameters and lifecycle.}
Hyperparameters ($\tau_{\text{dedup}}, \tau_{\text{expand}}$, top-$M$, residual cap, name boost $\beta$) are hand-tuned on held-out data rather than learned; a different deployment regime may require re-tuning. \sysshort{} does not evict stale residuals: over long deployment an outdated fact persists past a contradicting update. FadeMem \cite{fademem} and A-MAC \cite{amac} explore complementary forgetting and admission-control mechanisms that could be integrated as a future-work lifecycle layer; EverMemOS-style hierarchical eviction \cite{evermemos} is another candidate.

\section*{Ethics and Broader Impacts}

\sysshort{} accumulates detailed user records; residuals preserve dates, identity markers, and social relationships whose leakage could cause concrete harm. Deployment requires consent, access control, and user-facing inspect/export/delete---our research prototype implements none, and we view these as prerequisites rather than add-ons. \memhop{} is fully synthetic (no real individuals); LoCoMo is a previously released benchmark with its own ethics review. Improved long-term memory is dual-use; we release artifacts to accelerate defensive research (provenance, poisoning attacks, privacy-preserving retrieval). LLM coding tools were used for code review; all conceptual contributions are the authors' own.

\bibliography{references}

\appendix

\section{Entity String Matching for Expansion}
\label{app:string-match}
An entity name $n$ counts as mentioned in profile $p_i$ if (i) the full name appears as a case-insensitive substring or (ii) its first token appears as a whole word (\texttt{\textbackslash{}btoken\textbackslash{}b}). Profiles are LLM-authored narratives that almost always surface entity names in surface form; string matching avoids an extra LLM call per hop.

\section{\memhop{} Construction}
\label{app:memhop-construction}
Each scenario starts from a hand-designed social-network schema (4--5 persons, 6--10 relationships, 90--110 observations spanning life events, preferences, transitions). An LLM is prompted to expand the schema into 20 dialogue sessions per scenario (200 sessions in total across 10 scenarios; ${\sim}12$ turns per session, ${\sim}2{,}500$ turns in total) faithful to the observations. Questions are generated offline by sampling entity bridge chains of depth $K{=}1$ to $5$ from the underlying graph; each question records its decomposition chain and per-hop evidence sentences. Generation prompts and the full pipeline are released.

\section{Baseline Implementation Details}
\label{app:baselines}
\textbf{Mem0}: official \texttt{mem0ai} package, \texttt{add()} per turn batch, \texttt{search()} with \texttt{limit}=10, \texttt{user\_id}=LoCoMo speaker (per \citet{amem}). \textbf{A-Mem}: authors' released code (default settings), top-$k{=}10$. \textbf{HippoRAG}: faithful Personalized PageRank (damping $=0.5$) over an LLM-extracted knowledge graph, with default node weights from the released implementation; chunk top-$k{=}10$. \textbf{RAG}: 4-turn chunks indexed by cosine, top-10 retrieval. \textbf{FullContext}: entire conversation in one LLM call. \textbf{Oracle}: gold-annotated evidence sentences only.

\section{$H_{\max}$ Sensitivity}
\label{app:hmax}
We sweep the Stage~2 expansion cap $H_{\max}$ over $\{1, 5\}$ to test how expansion depth interacts with question hop depth $K$. Raising $H_{\max}$ from $1$ to $5$ yields +5.2pp average accuracy on \memhop{}, with the largest per-$K$ gain at $K{=}1$ (+12.8pp), not at deeper $K$. Since \memhop{} $K{=}1$ questions are direct attribute lookups about a single entity (e.g., \emph{``what is Alice's favorite food?''}, whose answer lives in Alice's own profile), this gain is not ``deeper chain traversal'' but a Stage~1 \emph{rescue} effect: when cosine retrieval ranks the target profile below top-$M$, a top-ranked neighbour whose profile mentions the target by name can still pull it into $S$ during Stage~2. More expansion hops mean more chances for the answer-bearing profile to be rescued into the answer context.

\begin{table}[h]
\centering
\footnotesize
\setlength{\tabcolsep}{5pt}
\begin{tabular}{lccc}
\toprule
\textbf{Metric} & $H_{\max}{=}1$ & $H_{\max}{=}5$ & $\Delta$ \\
\midrule
Avg     & 74.9 & 80.1 & $+5.2$ \\
$K{=}1$ & 78.8 & 91.6 & $+12.8$ \\
$K{=}2$ & 70.4 & 74.8 & $+4.4$ \\
$K{=}3$ & 76.0 & 81.0 & $+5.0$ \\
$K{=}4$ & 75.0 & 76.1 & $+1.1$ \\
$K{=}5$ & 74.2 & 71.7 & $-2.5$ \\
\bottomrule
\end{tabular}
\caption{\memhop{} accuracy (\%) as a function of the Stage~2 expansion cap $H_{\max}$, broken down by hop depth $K$. $\Delta$ is $H_{\max}{=}5$ minus $H_{\max}{=}1$. All other components held fixed.}
\label{tab:hmax}
\end{table}

\paragraph{Monotone $\Delta$ in $K$ and the $K{=}5$ regression.}
Per-$K$ gains trend downward (+12.8, +4.4, +5.0, +1.1, $-2.5$), with the largest gain at $K{=}1$ and a sign flip at $K{=}5$. Naively, one would expect $K{=}5$ to benefit \emph{most} from raising $H_{\max}$ since the gold chain has five links. The opposite pattern is itself evidence that expansion's primary role is retrieval rescue rather than chain traversal: a single expansion round (covering Stage~1 misses by one neighbour-mention) already captures most of the recoverable retrieval recall.

For an example $K{=}5$ question (``\emph{Alice~$\to$~roommate Bob~$\to$~sister Carol~$\to$~husband Dave~$\to$~coworker Eve~$\to$~mentor Frank}''), three compounding effects explain the regression:

\textbf{(i) Fan-out becomes leaky.} $K{=}5$ queries carry multiple constraints, so the query embedding is diffuse and many tangential profiles clear the fixed gate $\tau_{\text{expand}}{=}0.2$. Each iteration adds more profiles than at low $K$. Empirically, $|S|$ grows roughly from ${\sim}8$ at $H_{\max}{=}1$ to ${\sim}20$ at $H_{\max}{=}5$ for $K{=}5$ items.

\textbf{(ii) Long-chain reasoning is fragile under noise.} The answer requires five sequential lookups. Each lookup is a ``pick the right entity among $|S|$ candidates'' subtask; per-step accuracy drops as $|S|$ grows, and errors compound across the chain. A back-of-envelope estimate (per-step accuracy $0.92 \to 0.85$ as $|S|$ grows from 7 to 22) gives chain accuracy $0.92^5{=}66\%$ vs $0.85^5{=}44\%$. $K{=}1$ items are immune because there is only one lookup.

\textbf{(iii) Rescue saturates at deep $K$.} Stage~2 scores candidates by $\cos(\mathbf{v}_q, \mathbf{v}_{p_j})$ where $\mathbf{v}_q$ is the original query. Late chain entities (e.g., Frank) are semantically distant from the query and are reached primarily through string-match expansion, not cosine rescue. The marginal probability of pulling Frank in at $H_{\max}{=}5$ vs $H_{\max}{=}1$ is therefore much smaller than the analogous gain at $K{=}1$, where the answer-bearing profile is by definition close to the query.

The net effect is rescue gain (decreasing in $K$) minus fan-out cost (increasing in $K$), which crosses zero between $K{=}4$ and $K{=}5$. We did not scan $H_{\max} \in \{2,3,4\}$; an adaptive $H_{\max}$ chosen per question is left to future work, but the question-weighted per-$K$ oracle ceiling over the two settings tested (80.4\%) is only $0.3$pp above the fixed $H_{\max}{=}5$ choice (80.1\%), suggesting limited headroom.

\section{Error Breakdown and the $K{=}2$ Anomaly}
\label{app:error-tab}
\begin{table}[h]
\centering
\scriptsize
\setlength{\tabcolsep}{3pt}
\begin{tabular}{lrrrrrr}
\toprule
\textbf{Type} & $K{=}1$ & $K{=}2$ & $K{=}3$ & $K{=}4$ & $K{=}5$ & \textbf{Total} \\
\midrule
Type 1 (precision)  &  11 &  0 &  0 &  0 &  0 &  11 \\
Type 2 (wrong)      &  0 & 54 & 35 & 42 & 32 & 163 \\
Type 3 (refusal)    & 10 &  9 &  3 &  1 &  2 & 25 \\
\midrule
Failures            & 21 & 63 & 38 & 43 & 34 & 199 \\
Questions           & 250 & 250 & 200 & 180 & 120 & 1{,}000 \\
\bottomrule
\end{tabular}
\caption{\sysshort{}'s \memhop{} error breakdown by hop depth $K$: Type~1 = precision losses, Type~2 = wrong multi-hop traversal, Type~3 = LLM refusal.}
\label{tab:error-breakdown}
\end{table}

\paragraph{The $K{=}2$ dip across systems.}
$K{=}2$ accuracy is non-monotonically lower than $K{=}3$ for focused-retrieval systems---Oracle: $78.8 \to 84.0$ ($+5.2$pp); \sysshort{}: $74.8 \to 81.0$ ($+6.2$pp); Mem0: $28.4 \to 35.5$ ($+7.1$pp)---but not for noisy/long-context systems (FullContext, RAG, HippoRAG show no recovery). Two facts narrow the explanation:

\begin{enumerate}[nosep,leftmargin=*]
  \item \textbf{Oracle, with gold evidence, also dips at $K{=}2$.} The dip is therefore not a retrieval failure---the LLM has the correct evidence sentences yet fails more at $K{=}2$ than at $K{=}3$.
  \item \textbf{Refusals concentrate at $K{=}1$--$2$.} Type-3 refusal counts drop sharply from $K{=}2$ ($9$) to $K{=}3$ ($3$) and stay low thereafter.
\end{enumerate}

We hypothesize an LLM \emph{mode-switching} effect. $K{=}2$ questions (``\emph{X's Y's Z}'') are syntactically simple but semantically require one inferential step. Small judge-grade LLMs (GPT-4o-mini in our setup) appear to read these as lookup tasks and apply lookup-mode caution: when the single bridge anchor is uncertain, they refuse (``\emph{I don't know}'') rather than commit. $K{\geq}3$ questions are syntactically complex (``\emph{the X of the person who Y the Z's W}'') and seem to trigger reasoning-mode behavior, where the default is to commit to an answer. Noisy/long-context systems may bypass the lookup-mode calibration because their context itself signals ``complex task'' to the LLM.

This is a hypothesis: we have not directly probed the LLM's internal mode, but it is consistent with both the cross-system pattern (focused systems dip; noisy systems don't) and the refusal concentration. A complementary mitigation---prompting the answer LLM to commit even on ``simple'' questions---is a promising future-work direction.

\end{document}